\documentclass[letterpaper]{article} 
\usepackage{aaai2026}  
\usepackage{times}  
\usepackage{helvet}  
\usepackage{courier}  
\usepackage[hyphens]{url}  
\usepackage{graphicx} 
\usepackage{natbib}  
\usepackage{caption} 
\usepackage{algorithm}
\usepackage{algorithmic}

\usepackage{newfloat}
\usepackage{listings}
\DeclareCaptionStyle{ruled}{labelfont=normalfont,labelsep=colon,strut=off} 
\floatstyle{ruled}
\newfloat{listing}{tb}{lst}{}
\floatname{listing}{Listing}
\usepackage[utf8]{inputenc}
\usepackage{caption}
\usepackage{amsmath}
\usepackage{amsthm}
\usepackage{booktabs}
\usepackage{algorithm}
\usepackage{algorithmic}
\usepackage[switch]{lineno}

\usepackage{amssymb}
\usepackage{tabularx}
\usepackage{multirow}
\usepackage{multicol}
\usepackage{subcaption}
\usepackage{xcolor}
\usepackage{pifont}
\usepackage{enumitem}

\nocopyright

\title{RISE-T2V: Rephrasing and Injecting Semantics with LLM for Expansive Text-to-Video Generation}
\author{
    Xiangjun Zhang\textsuperscript{\rm 1}, Litong Gong\textsuperscript{\rm 2}, Yinglin Zheng\textsuperscript{\rm 1}, Yansong Liu\textsuperscript{\rm 1}, Wentao Jiang\textsuperscript{\rm 2}, Mingyi Xu\textsuperscript{\rm 1}, Biao Wang\textsuperscript{\rm 2}, Tiezheng Ge\textsuperscript{\rm 2}, Ming Zeng\textsuperscript{\rm 1}
}
\affiliations{
    \textsuperscript{\rm 1} Xiamen University, Xiamen, China\\
    \textsuperscript{\rm 2}  Alibaba Group, Beijing, China

    \{zhangxiangjun@stu., zhengyinglin@stu., liuyansong@stu., xumingyi2020@stu., zengming@\}xmu.edu.cn, \{gonglitong.glt, winter.jwt, eric.wb, tiezheng.gtz\}@alibaba-inc.com
}

\usepackage{bibentry}

\begin{document}

\maketitle

\begin{abstract}
    Most text-to-video (T2V) diffusion models depend on pre-trained text encoders for semantic alignment, yet they often fail to maintain video quality when provided with concise prompts rather than well-designed ones. The primary issue lies in their limited textual semantics understanding. Moreover, these text encoders cannot rephrase prompts online to better align with user intentions, which limits both the scalability and usability of the models. To address these challenges, we introduce RISE-T2V, which uniquely integrates the processes of prompt rephrasing and semantic feature extraction into a single and seamless step instead of two separate steps. RISE-T2V is universal and can be applied to various pre-trained LLMs and video diffusion models(VDMs), significantly enhancing their capabilities for T2V tasks. We propose an innovative module called the Rephrasing Adapter, enabling diffusion models to utilize text hidden states during the next token prediction of the LLM as a condition for video generation. By employing a Rephrasing Adapter, the video generation model can implicitly rephrase basic prompts into more comprehensive representations that better match the user's intent. Furthermore, we leverage the powerful capabilities of LLMs to enable video generation models to accomplish a broader range of T2V tasks. Extensive experiments demonstrate that RISE-T2V is a versatile framework applicable to different video diffusion model architectures, significantly enhancing the ability of T2V models to generate high-quality videos that align with user intent. Visual results are available on the webpage at {\textit{https://rise-t2v.github.io}}.
\end{abstract}

\section{Introduction}
Recently, text-to-image (T2I) generation based on diffusion models~\cite{ho2020denoising,song2022denoising,Rombach_2022_CVPR} has achieved significant improvements. Advanced T2I models~\cite{Rombach_2022_CVPR,podell2023sdxl,chen2023pixartalpha} are trained on large-scale multimodal datasets~\cite{schuhmann2021laion400mopendatasetclipfiltered,schuhmann2022laion}, which can generate a variety of realistic images based on a given textual prompt in an end-to-end manner. In addition, text-to-video (T2V) generation such as CogVideoX~\cite{yang2024cogvideoxtexttovideodiffusionmodels} and AnimateDiff~\cite{guo2024animatediff} has also recently achieved significant enhancements to generate high-quality videos based on provided text.

\begin{figure}[tbp]
    \centering
    \includegraphics[width=1\linewidth]{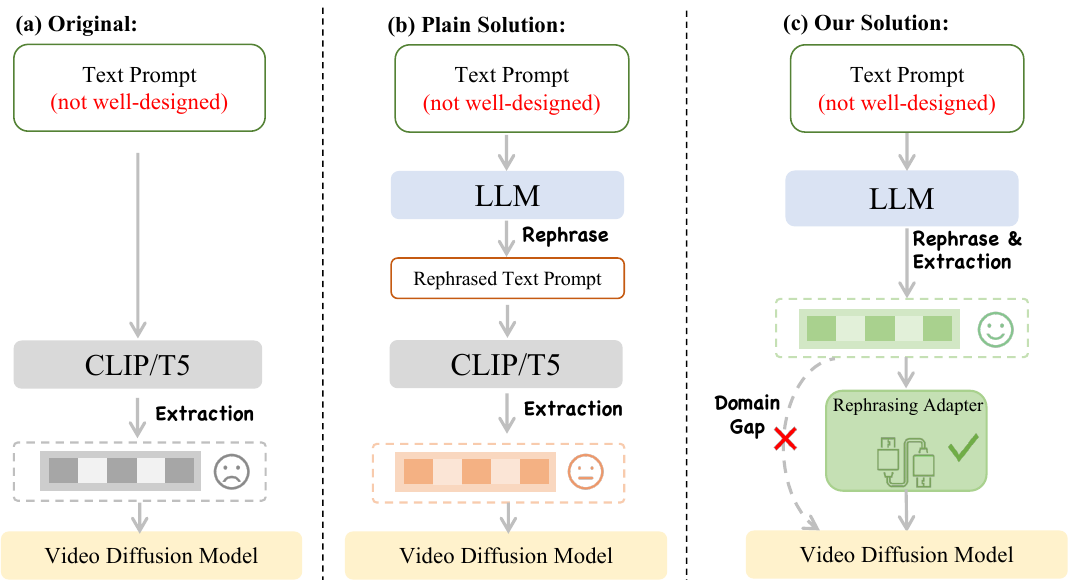} 
    \caption{High-level pipeline of our method. (a). Directly using CLIP/T5 as a feature extractor. (b). Using LLM as prompt rewriter and feeding text into CLIP/T5 for feature extraction. (c). Our method combines prompt rephrasing with semantic feature extraction in a seamless process. The proposed Rephrasing Adapter bridges the gap between LLM-rephrasing feature and pre-trained diffusion model.\vspace{-1em}}
    \label{fig:pipeline}
    \vspace{-1em}
\end{figure}
In existing open-source T2V models, the quality of generated videos largely depends on the provided textual prompts. Videos generated from simple prompts are usually inferior to those created with well-crafted prompts (see Figure~\ref{fig:pipeline}(a)). Well-designed prompts tend to produce higher-quality videos that better reflect user intent. However, it is often inconvenient for users to compose such prompts, so they typically opt for simpler descriptions for efficiency.
Currently, LLM-based generative models have been extensively explored in both text-to-image (T2I) and text-to-video (T2V) domains, which can be categorized into two major approaches: LLM-Based Encoders and Two-stage Rephrasing. 
LLM-Based Encoders (such as CogVideoX~\cite{yang2024cogvideoxtexttovideodiffusionmodels}) employ large language models (LLMs) as the text encoder. Two-stage Rephrasing methods (see Figure~\ref{fig:pipeline}(b)) use LLMs to rephrase user prompts, and then pre-trained encoders like CLIP~\cite{radford2021learning} or T5~\cite{raffel2023exploringlimitstransferlearning} to encode the rephrased text. 
As summarized in Table~\ref{tab:core_comparison}, 
LLM-Based Encoders lack rephrasing ability, while two-stage rephrasing methods introduce redundant steps and lead to semantic gaps between separated steps.

To address these issues, as illustrated in Figure~\ref{fig:pipeline}(c), we propose a one-step RISE-T2V approach, which leverages LLMs for both prompt rephrasing and semantic feature extraction, making the method more flexible and efficient. To achieve this, we design a novel Rephrasing Adapter (RA) that enables diffusion models to directly use LLM-generated text encodings as conditions for T2V models.
During inference, the LLM rephrases the text prompt and extracts the hidden states of next token prediction as a conditional injection.  As depicted in Figure~\ref{fig:pipeline}(c), we use RA to bridge between the predicted hidden states of the LLM and the pre-trained VDM, injecting these rephrased hidden states into the diffusion model. The video diffusion model generates videos based on the rephrased text encodings. We achieve rapid semantic adaptation of pre-trained models through chat adaptation and motion adaptation training phases, exhibiting better semantic understanding and motion fluency.
\begin{table}[t!]
  \centering
  \setlength{\tabcolsep}{0.4mm}
  \begin{tabular}{ccc}
    \toprule
    \textbf{Methods} & \textbf{Prompt Rephrase} & \textbf{Unified Process} \\
    \midrule
    LLM-Based Encoders & \ding{55} & \ding{51} \\
    Two-stage Rephrasing & \ding{51} & \ding{55} \\
    \textbf{Ours} & \ding{51} & \ding{51} \\
    \bottomrule
  \end{tabular}
  \vspace{-1em}
  \caption{Capability comparison of different T2V frameworks. The checkmarks indicate support for each feature.\vspace{-1em}}
  \label{tab:core_comparison}
\end{table}

By integrating LLMs with T2V models, users can leverage rephrasing instructions to convert text prompts into more detailed and intent-aligned text encodings. This integration enhances the scalability of T2V tasks, making them applicable to a wider range of scenarios, including but not limited to dense text-encoded video generation, multi-scene text-encoded video generation, and multilingual text-encoded video generation, all enabled by different rephrasing instructions. Experimental results demonstrate that combining LLMs with diffusion models can significantly improve T2V generation performance, thanks to the LLM’s extensive world knowledge, planning ability, and causal reasoning capabilities.

In summary, our contributions are as follows:
\begin{itemize}
    \item We introduce a new approach named RISE-T2V by integrating LLMs into existing T2V diffusion models, significantly enhancing the generative capabilities of pre-trained T2V models.
    \item We develop the Rephrasing Adapter to combine the processes of prompt rephrasing and semantic feature extraction into a single, seamless step instead of two separate steps. 
    \item Coupled with the LLM's powerful capabilities, we can create user-satisfying and high-quality videos from simple prompts. Our experiments have demonstrated that our method enables generative models to complete a wider variety of T2V tasks, thereby enhancing the user experience.
\end{itemize}

\section{Related Work}
\subsection{Text-to-Video Diffusion}
Text-to-video generation involves the creation of realistic videos from natural language descriptions. Recent advancements~\cite{ho2022video,ho2022imagen,wang2023modelscope,ma2024latte,gong2024atomovideohighfidelityimagetovideo} have seen the use of diffusion models~\cite{ho2020denoising,song2022denoising,Rombach_2022_CVPR} in this domain. These work can be divided into two main categories, i.e. UNet-based~\cite{Rombach_2022_CVPR} and DiT-based(Diffusion Transformer)~\cite{peebles2023scalablediffusionmodelstransformers}. 
For UNet-based methods, VDM~\cite{ho2022video} extends the conventional image diffusion architecture to integrate both image and video data, addressing T2V generation tasks. ModelScope T2V~\cite{wang2023modelscope} introduces spatio-temporal blocks to model temporal dependencies, ensuring the generation of consistent frames and smooth motion transitions. In recent research, numerous studies~\cite{khachatryan2023text2videozero,wu2023tuneavideo,guo2024animatediff,blattmann2023align} have sought to leverage the power of T2I models to improve video generation quality. AnimateDiff~\cite{guo2024animatediff} designs a plug-and-play motion module, which can drive various personalized T2I models to generate animations. 
For DiT-based methods, CogVideoX~\cite{yang2024cogvideoxtexttovideodiffusionmodels} presents an innovative DiT-based video diffusion model, which uses 3D full attention to effectively capture the spatiotemporal distribution in videos. 
While previous studies have achieved temporally consistent and high-fidelity video generation, the quality of the result depends on the prompt. 
In our study, we utilize LLMs alongside Rephrasing Adapter to implicitly rephrase the input text prompt, leading to an overall improvement in video quality.

\subsection{LLM-Enhanced Video Generation}
To leverage the capabilities of LLMs in generative tasks, some methods~\cite{hu2024ella,zhao2024bridging,liu2024llm4genleveragingsemanticrepresentation,Mimir2025,LDGen2025} employ LLMs as the text encoder for diffusion models, thereby enhancing the model's ability to understand prompts.
To employ LLMs in video generation, previous research~\cite{huang2024free,hong2024direct2v,bansal2024talctimealignedcaptionsmultiscene,lin2024videodirectorgptconsistentmultiscenevideo,lian2024llmgroundeddiffusionenhancingprompt} utilize causal reasoning abilities of LLMs to create prompts that align closely with user intentions. Free-Bloom~\cite{huang2024free} and DirecT2V~\cite{hong2024direct2v} use LLMs to convert text prompts into a series of narrative events, which unfold over time with frame-by-frame prompts, ensuring semantic consistent video generation. 
The previous methods either use the LLM solely as a text encoder without leveraging its powerful generative capabilities or employ the LLM to generate extra text information that is explicitly injected into the generative model, relying on another text encoder in a two-step process. 
Our approach employs the RA to perform prompt rephrasing and semantic feature extraction in one step, directly utilizing the original predicted hidden states of the LLM and achieving superior results.
\begin{figure*}[ht]
    \centering
    \vspace{-0.4in}
    \includegraphics[width=0.9\textwidth]{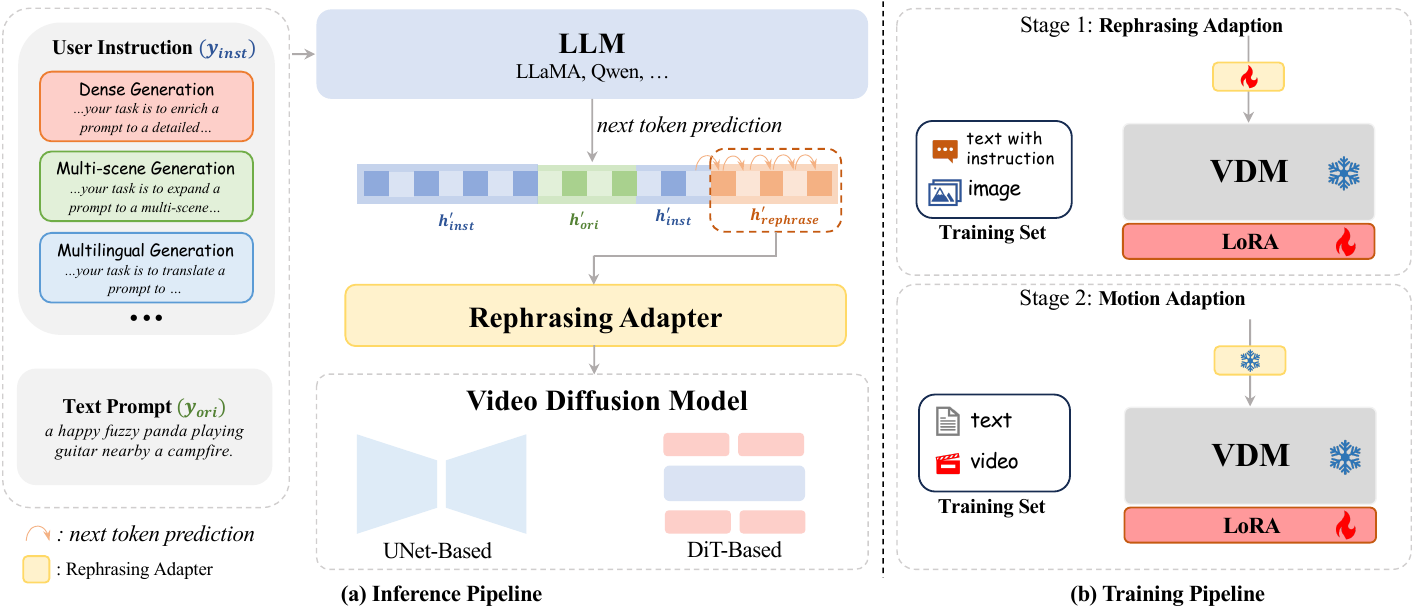} 
    \vspace{-0.5em}
    \caption{Overview. (a). The inference pipeline of RISE-T2V. The Rephrasing Adapter can integrate with various LLMs and diffusion models. It enables diffusion models to utilize the text hidden states from the LLM's next token prediction, serving as a condition for video generation. (b).The training scheme for RISE-T2V. In stage 1, we train the RA to adapt the rephrased text encodings to the diffusion model. In stage 2, we train the model on videos to achieve motion adaptation.}
    \label{fig:method}
    \vspace{-1em}
\end{figure*}
\section{Method}
\subsection{Overview of RISE-T2V}
Figure~\ref{fig:method}(a) illustrates the inference pipeline of RISE-T2V. We begin by feeding the LLM a rephrasing instruction together with a text prompt. The LLM generates hidden states corresponding to the rephrased text via next token prediction. These generated hidden states are then passed through the Rephrasing Adapter and converted into text encodings suitable for the video diffusion model (VDM). Finally, this encoding is injected into the attention module of the VDM to guide video generation. Unlike previous approaches that only use the LLM as a static text encoder, adapting features derived from next token prediction in LLMs to diffusion models is not straightforward. Therefore, we specifically design a Rephrasing Adapter and construct dedicated chat training data to facilitate adaptation through two stages of learning. Through these steps, our method can generate high-quality videos that are more consistent with user instructions, especially in terms of richer details and smoother motion.
RISE-T2V is a versatile framework that can seamlessly integrate different pre-trained language models, such as LLaMA~\cite{touvron2023llama2openfoundation} or Qwen~\cite{yang2024qwen2technicalreport}, and video generation models, such as UNet-based or DiT-based models. This unified single-step approach, which combines prompt rephrasing and semantic injection, enables video generation that more closely matches user intent and supports diverse, high-quality effects.

\subsection{Seamless Prompt Rephrasing and Semantic Feature Extraction with LLMs.}
\label{sec:rewrite}

Given a text prompt $y_{ori}$, and a LLM $\Phi$, existing methods~\cite{zhao2024bridging,hu2024ella} only use LLM as simple text feature extractor, using $h_{ori} = E_\Phi(y_{ori})$ for text representation, where $E_\Phi$ refers to the final hidden states obtained during the encoding phase with the $\Phi$. However, the limited information in $y_{ori}$ often results in unsatisfactory representations that do not fully leverage the generative capabilities of LLMs. To enhance text conditioning, we propose using instructions to guide the LLM in rephrasing the original text instead of directly encoding $y_{ori}$. We first use a rephrasing instruction template $y_{inst}$ to provide specific guidance for enhancing $y_{ori}$. We then employ the LLM $\Phi$ to perform text generation with $y_{inst}$ and $y_{ori}$ via next token prediction, resulting in the rephrased text $y_{rephrase}$. As characteristic of LLMs with GPT architecture, this process not only generates expanded text, but also extracts features of the entire sentence(including $y_{inst}$, $y_{ori}$, $y_{rephrase}$), which can be expressed as follows:
\begin{equation}
    E_\Phi(y_{inst}+y_{ori}+y_{rephrase}), y_{rephrase} = G_\Phi(y_{inst}+y_{ori}),
    \label{eq:rewrite}
\end{equation}
where $G_\Phi$ denotes the text generation process using the LLM $\Phi$.

A straightforward approach is to employ the LLM again as text encoder, encoding $y_{rephrase}$, thereby generating $h_{rephrase} = E_\Phi(y_{rephrase})$ as the rephrased hidden states for subsequent processing steps. However, as shown in Eq.\ref{eq:rewrite}, during inference, the hidden states of $y_{rephrase}$ are generated through next token prediction by the LLM using $y_{inst}$ and $y_{ori}$. Therefore, the ``actual" $y_{rephrase}$ hidden states for inference can be represented as follows:
\begin{equation}
    h_{inst}^{\prime}, h_{ori}^{\prime}, h_{rephrase}^{\prime} = E_\Phi(y_{inst}+y_{ori}+y_{rephrase})
\label{eq:decompose}
\end{equation}
Owing to the inherent properties of LLM, the hidden states are token-level aligned with the input text, thus can be easily divided into three components $h_{inst}^{\prime}$, $h_{ori}^{\prime}$ and $h_{rephrase}^{\prime}$. Since $h_{rephrase}^{\prime}$ is also the hidden state of $y_{rephrase}$, it serves as a rephrased text representation, which we refer to as the \textit{rephrasing feature}. In contrast, $h_{rephrase}$ is termed \textit{encoded feature}. It's important to note that rephrasing feature and encoded feature differ slightly due to the influence of additional text components, particularly user instructions, creating a domain discrepancy. If the RA is trained to accept encoded features as input, while inference time, it accepts the rephrasing features as input, there is a discrepancy between them, it often results in artifacts like blurriness and color distortion, as demonstrated by the panda case in the Figure~\ref{fig:chat adapter}(b).

\subsection{Adapting Rephrasing Feature for Video Generation with Rephrasing Adapter.}
\label{sec:training}
\begin{figure}[t]
    \centering
    \includegraphics[width=0.9\linewidth]{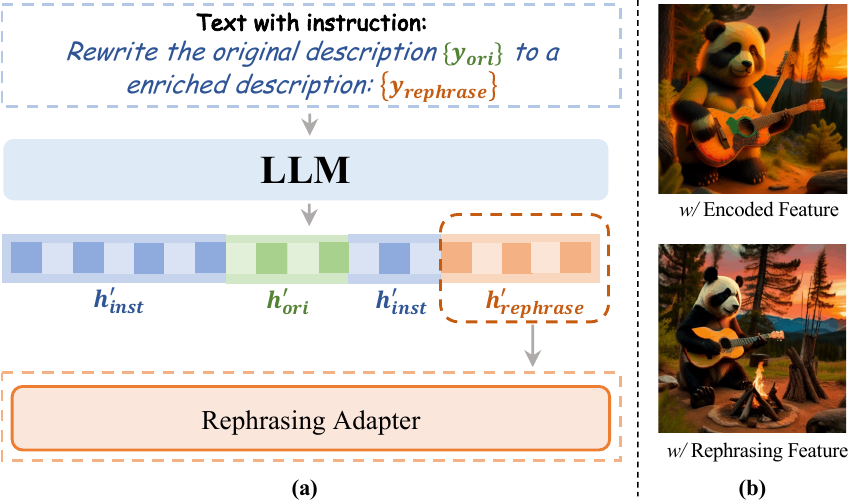} 
    \caption{Illustration of RA. (a).The training data for the RA is constructed by combining the user instruction $y_{inst}$, original text prompt $y_{ori}$, and enriched text prompt $y_{rephrase}$ are combined into a unified input text, encoded by the LLM, and the hidden states $h_{rephrase}^{\prime}$ are extracted for the training. (b).Visual Comparison: Rephrasing Feature vs. Encoded Feature. The lower row image is clearer and more aesthetically pleasing.}
    \label{fig:chat adapter}
    \vspace{-1em}
\end{figure}

To correctly and effectively train RA with the rephrasing feature, we constructed a dataset with the aligned quadruple annotation ($y_{inst}$, $y_{ori}$, $y_{rephrase}$, $x_0$), where $y_{rephrase}$ is the rephrased text produced by multi-modal LLM~\cite{bai2023qwenvlversatilevisionlanguagemodel} given image-text pair ($x_0$,$y_{ori}$). Therefore, in the training stage, the rephrasing feature can be obtained efficiently using Eq.\ref{eq:decompose} as shown in Figure~\ref{fig:chat adapter}(a). Specifically, as outlined in Figure~\ref{fig:chat adapter}(a), 
we extract $h_{rephrase}^\prime$ and employed it for further training of the LLM \textbf{R}ephrasing \textbf{A}dapter $f_{RA}$, and the rephrased text encodings $c$ can be written as:
    \begin{align} 
        c = f_{RA}(h_{rephrase}^\prime)
    \label{eq:chatadapter}
    \end{align}
We implement $f_{RA}$ using stacked feed-forward layers. 
To make full use of the existing vast image-text pair data and reduce manual labeling workload, we designed a two-phased training scheme which is depicted in Figure~\ref{fig:method}(b). 
In stage 1, we freeze the pretrained LLM and diffusion model, and train the RA and LoRA~\cite{hu2021lora} injected to VDM on the text-image quadruple data. In stage 2, to learn motion priors, we freeze RA and only finetune the trained LoRA on text-video pair data.

\subsection{Applications}
The Rephrasing Adapter connects LLM with diffusion models, allowing for the transformation of input text prompts into higher-quality, user-aligned formats based on user instructions. 
The refined text encodings directly facilitate video generation, extending the application range beyond standard T2V models. This includes capabilities such as instruction-optimized dense text encoding generation, multi-scene text encoding generation, and multilingual text encoding generation. 

\paragraph{Dense Text Encoding Generation.} Videos generated from simple prompts often lack the quality achieved with more detailed, longer prompts. However, crafting such descriptions can be inconvenient for users. Leveraging LLMs with enrichment instructions to provide richer text encodings for video generation is the most direct application of our method, significantly enhancing the aesthetic quality and alignment of the output videos.

\paragraph{Multi-scene Text Encoding Generation.} Current T2V models often struggle to generate temporally aligned multi-scene videos. Figure~\ref{fig:multi_scene} illustrates that AnimateDiff fails to generate temporal transitions, such as "from holding its arms down to raising them" and "from idling to trotting". We aim to take a brief text prompt and leverage the advanced text generation and reasoning capabilities of an LLM to construct a multi-scene script. The serialized text encodings from this script are then used for video generation, thereby achieving temporally coherent multi-scene video generation.

\paragraph{Multilingual Text Encoding Generation.} Leveraging the inherent multilingual capabilities of LLMs, our framework can process prompts in languages not present in the VDM's training data. It effectively converts foreign-language intent into aligned text encodings, enabling cross-lingual video generation without any model fine-tuning.

\section{Experiments}
\subsection{Implementation Details}
RISE-T2V can be applied to any decoder-only LLM and any LoRA-compatible VDM. For demonstration purposes, we proposed RISE-Animatediff, which boosts UNet-based VDM Animatediff\cite{guo2024animatediff} with LLaMA2-Chat 7B~\cite{touvron2023llama2openfoundation} under the RISE-T2V framework.  
All content is processed at 512 resolution, with video clips having 16 frames. RA comprises two feedforward layers. 
The RA's input dimensions match the rephrased hidden states, while its output dimensions align with the VDM's attention module.
In the first stage, we train the RA using 12 million pairs of text with instructions and images. The second stage focuses on learning motion priors from one million text and video pairs. More detailed information regarding the training process, hyperparameter settings, evaluation, and dataset construction can be found in the supplementary materials. We evaluate the effectiveness of our method on three expansive tasks derived from LLM capabilities. 

\subsection{Video Generation with Dense Text Encoding}
We evaluate the performance of dense text encodings in T2V tasks by comparing them with both baseline and other T2V models.
\subsubsection{Quantitative Results} We conducted a quantitative comparison using automatic metrics on the Prompt Suite per Category of VBench~\cite{huang2023vbench}, involving 800 prompts for eight distinct types: \textit{Animal, Architecture, Food, Human, Lifestyle, Plant, Scenery, and Vehicles}. Our evaluation focuses on three key aspects: aesthetic quality, motion smoothness, and text alignment, summarizing results into an overall score based on average rank (with lower numbers indicating better performance). 
As shown in Table~\ref{table:eval on baseline}, compared to the baseline method AnimateDiff (as depicted in Figure~\ref{fig:pipeline}(a)), we observed a comprehensive improvement across all metrics, particularly in terms of aesthetics and text alignment. This indicates that using the revised dense text encoding can effectively enhance the video quality. Next, we compare our method with AnimateDiff equipped with LLM (corresponding to Figure~\ref{fig:pipeline}(b)). The LLM transforms prompts from the VBench test set into more complex, detailed ones, enhancing text semantics. Quantitative results are in the second row of Table \ref{table:eval on baseline}. Our method exhibits superior aesthetics and smoothness. It is crucial to highlight that while AnimateDiff uses dense prompts directly (Figure 1(b)), involving two processes, our approach utilizes simpler prompts with less semantic information and avoids the need for an extra text encoder.

\begin{table}[tbp]
  \centering
  \setlength{\tabcolsep}{0.4mm}
  \footnotesize
    \begin{tabular}{c|c|ccc}
    \toprule
    Method & Input & Aesthetic↑ &  Motion↑  &   Text↑  \\
    \midrule
    AnimateDiff & Simple Prompt & 6.39  & \underline{0.983}  & 31.36  \\
    AnimateDiff$_\text{llm}$ & Dense Prompt & \underline{6.54}  & 0.981 & \textbf{32.69} \\
    RISE-AnimateDiff & Simple Prompt & \textbf{6.61} & \textbf{0.984} & \underline{31.95}   \\
    \bottomrule
    \end{tabular}%
    \vspace{-1em}
    \caption{\rm Quantitative comparison with the baseline models. The \textbf{bold} font represents the best result. \underline{Underlining} represents suboptimal result.}
  \label{table:eval on baseline}%
\end{table}%
\begin{table}[tbp]
  \centering
  \setlength{\tabcolsep}{0.4mm}
  \footnotesize
    \begin{tabular}{ccccc}
    \toprule
    Method & Aesthetic↑ & Motion↑ & Text↑ & Rank↓ \\
    \midrule
    Pika  & 6.11  & \textbf{0.996} & 29.65 & 4.33 \\
    Gen2  & \underline{6.45}  & \underline{0.995} & 31.35 & \underline{3.33} \\
    ModelScope & 4.94  & 0.967 & 31.61 & 6.67 \\
    Latte & 5.60  & 0.970 & \textbf{32.07} & 4.67 \\
    VideoCrafter2 & 6.00  & 0.984 & 31.86 & 3.67 \\
    CogVideoX-2B & 5.40 & 0.983 & 30.12 & 6.33 \\ 
    \midrule
    AnimateDiff & 6.39  & 0.983  & 31.36 & 4.33 \\
    RISE-AnimateDiff & \textbf{6.61} & 0.984 & \underline{31.95} & \textbf{2.00} \\
    \bottomrule
    \end{tabular}%
    \vspace{-1em}
    \caption{\rm Quantitative analysis of the evaluated T2V models.\vspace{-1em}}
    \label{table:eval on lck}%
    \vspace{-1em}
\end{table}%
We also compare our method with recent video generation models, including CogVideoX~\cite{yang2024cogvideoxtexttovideodiffusionmodels}, which utilizes a large language model (T5) as its text encoder, VideoCrafter2~\cite{Chen_2024_CVPR}, Latte~\cite{ma2024latte}, and ModelScope T2V~\cite{wang2023modelscope}, as well as commercial tools Gen-2~\cite{gen2} and Pika labs~\cite{pika}. 
As shown in Table~\ref{table:eval on lck}, our method achieves the highest scores in both aesthetic quality and average rank, and ranks second in text alignment. Overall, our method demonstrates comprehensive performance without any significant weaknesses, with even the lowest metric ranking third. Although motion smoothness is limited by the underlying model and text rephrasing offers limited improvement, our approach does not negatively impact smoothness compared to the baseline.

\begin{figure}[tbp]
    \begin{minipage}{0.2\textwidth}
        \includegraphics[width=\textwidth]{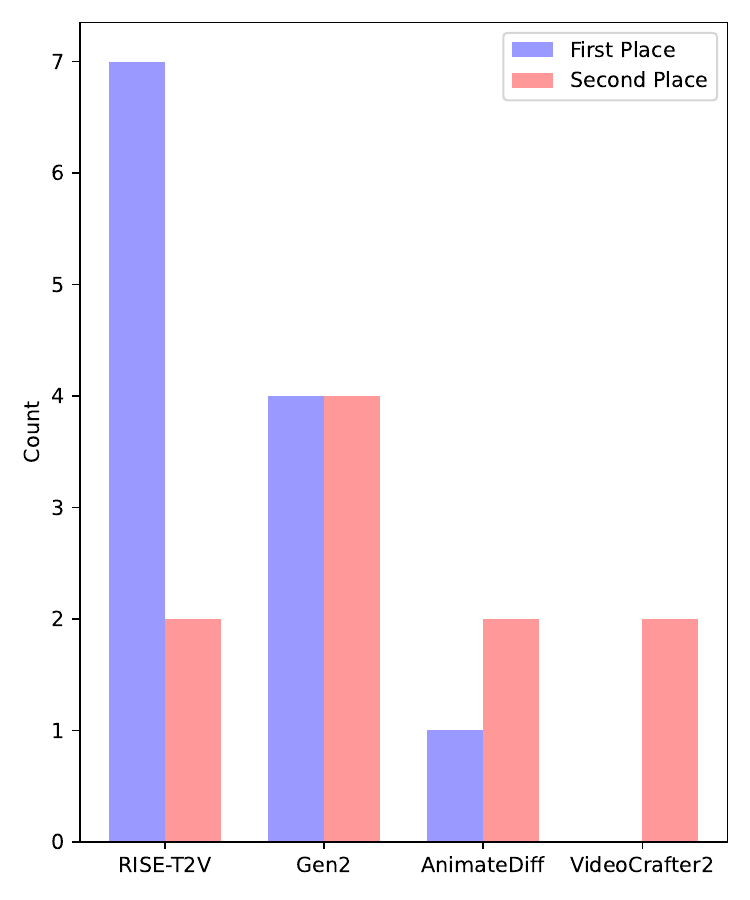} 
    \end{minipage}
    \begin{minipage}{0.3\textwidth}
        \hfill
        \includegraphics[width=\textwidth]{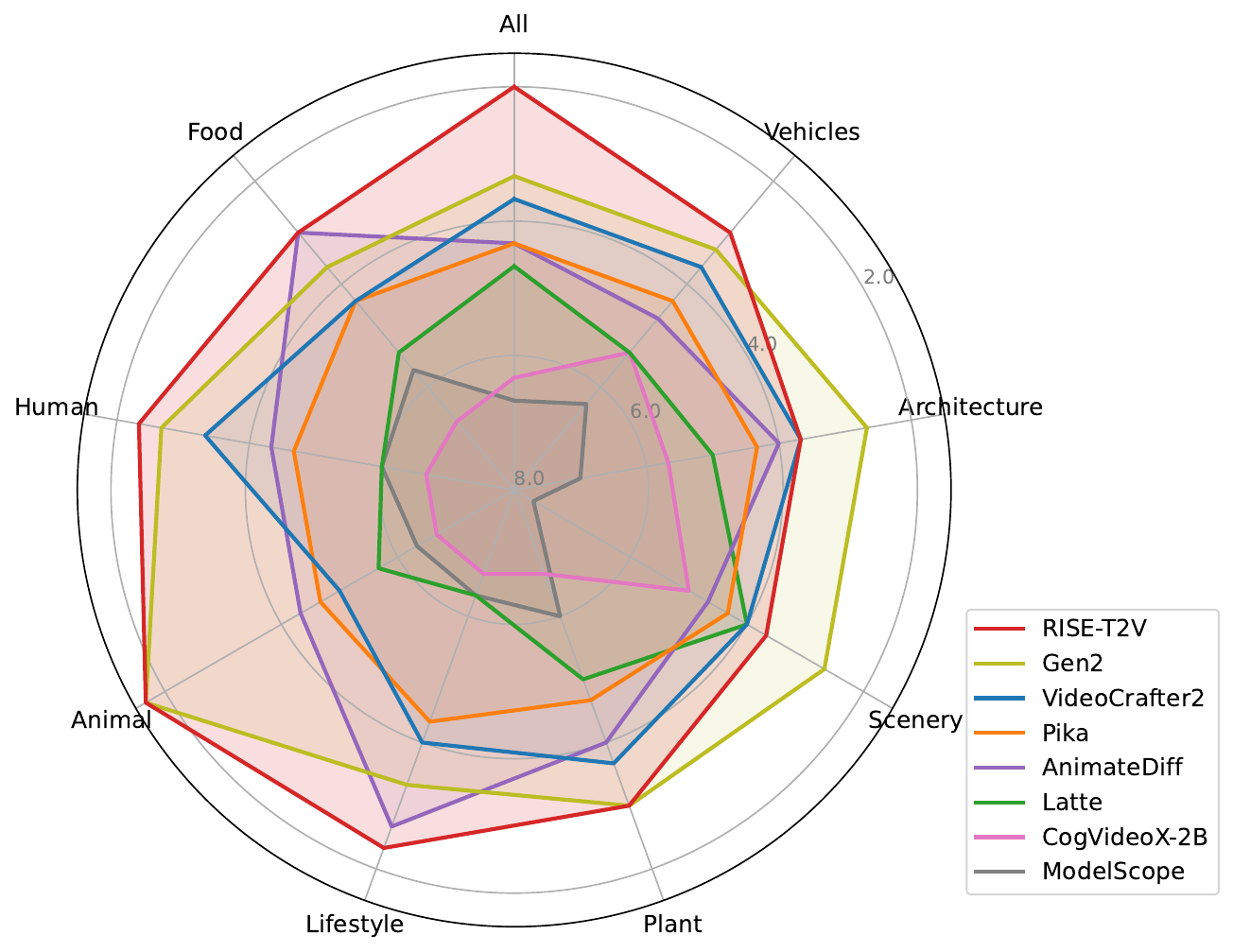} 
    \end{minipage}
    \begin{minipage}{0.45\linewidth}
        \subcaption{}
    \end{minipage}
    \quad
    \begin{minipage}{0.45\linewidth}
        \subcaption{}
    \end{minipage}
    \vspace{-1em}
    \caption{(a).The number of times RISE-AnimateDiff achieved first and second place across all subcategories (b).A comparison of the average ranks of RISE-AnimateDiff and other methods across 8 subcategories in the evaluation set.}
    \label{fig:radar}
    \vspace{-1em}
\end{figure}

For a comprehensive analysis, We conducted evaluations on 8 subcategories in the evaluation set and calculated average ranks for each. Figure~\ref{fig:radar} shows that RISE-T2V ranked higher than baseline methods, achieving first rank in six categories and second in two categories.
Besides, we conducted a human evaluation and selected three open-source methods that performed well on automatic metrics for comparison. The participants selected the best one (or multiple) results based on aesthetic quality, temporal quality, and text alignment, with percentages indicating selection proportions. 
In the appendix, we provide more detailed human evaluation settings. 
As shown in Table~\ref{tab:eval-on-lck-user-study}, our method achieves the highest voting rate across three aspects. 

\begin{table}[t]
  \centering
  \footnotesize
    \begin{tabular}{cccc}
    \toprule
    Method & Aesthetic↑ &  Temporal↑ & Text↑ \\
    \midrule
    Latte & 20.33\% & 18.67\% & 32.00\% \\
    VideoCrafter2 & 14.00\% & 16.33\% & 34.00\% \\
    AnimateDiff & 36.67\% & 33.33\% & 50.00\% \\
    \midrule
    RISE-AnimateDiff  & \textbf{49.00\%} & \textbf{54.00\%} & \textbf{52.66\%} \\
    \bottomrule
    \end{tabular}%
    \vspace{-1em}
    \caption{\rm User study results of the evaluated T2V models.\vspace{-1em}}
  \label{tab:eval-on-lck-user-study}%
  \vspace{-1em}
\end{table}%

\begin{figure*}[ht]
    \vspace{-0.3in}
    \centering
    \includegraphics[width=0.9\textwidth]{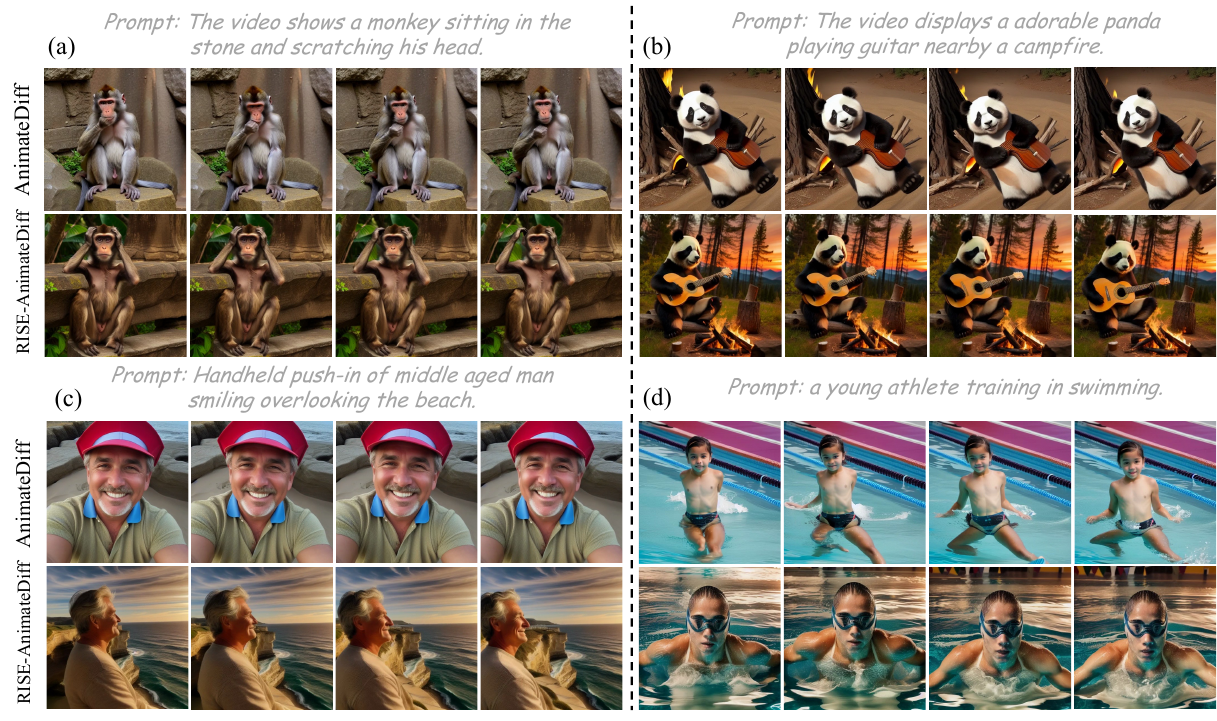}
    \vspace{-1em}
    \caption{Qualitative comparison. Our method can generate videos with high aesthetics and strong semantics alignment.\vspace{-0.5em}}
    \label{fig:lck_case}
    \vspace{-1em}
\end{figure*}
\subsubsection{Qualitative Results} Compared to our baseline methods, AnimateDiff, Figure~\ref{fig:lck_case} shows that our generated videos maintain high aesthetic quality and better alignment with prompts. For analysis, we convert the LLM-produced hidden states back into text. For example, in (a), the rephrased text describes: \textit{``In a lush, tropical environment, a monkey with grey and brown fur sits on a large stone, scratching his head with his right paw."} The LLM adds details like \textit{``lush, tropical environment"} and \textit{``grey and brown fur,"} enhancing the content's aesthetic appeal.

\subsection{Ablation Study}

Ablation experiments performed on dense text encoding validate the RA's effectiveness and RISE-T2V's generalization.
\paragraph{The Analysis of Rephrasing and Encoded Features.}
To evaluate the differences between rephrasing features and encoded features, we trained the RA with each feature as input respectively and compared the generated videos (see Figure~\ref{fig:chat adapter}(b)). While using encoded features maintains semantic correctness, the resulting visual clarity and aesthetic quality decrease significantly. This highlights the domain gap between the two types of features and underscores the importance of feature alignment to ensure high-quality generation.

\begin{figure}[tp]
    \centering
    \includegraphics[width=0.9\linewidth]{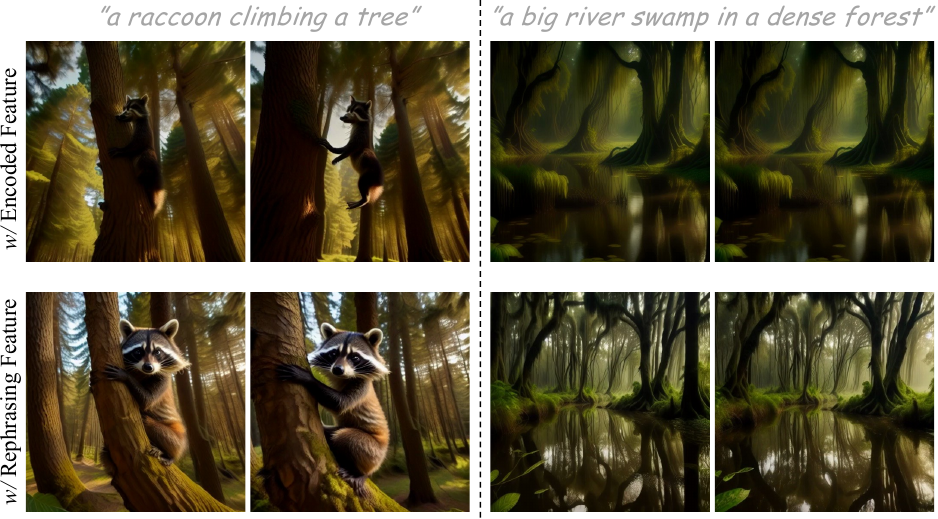}
    \vspace{-1em}
    \caption{Visual Comparison: Rephrasing Feature vs. Encoded Feature.\vspace{-0.5em}}
    \label{fig:ablation chat adapter}
    \vspace{-1em}            
\end{figure}
\begin{figure*}[htbp]
    \vspace{-0.1in}
    \centering
    \includegraphics[width=0.8\textwidth]{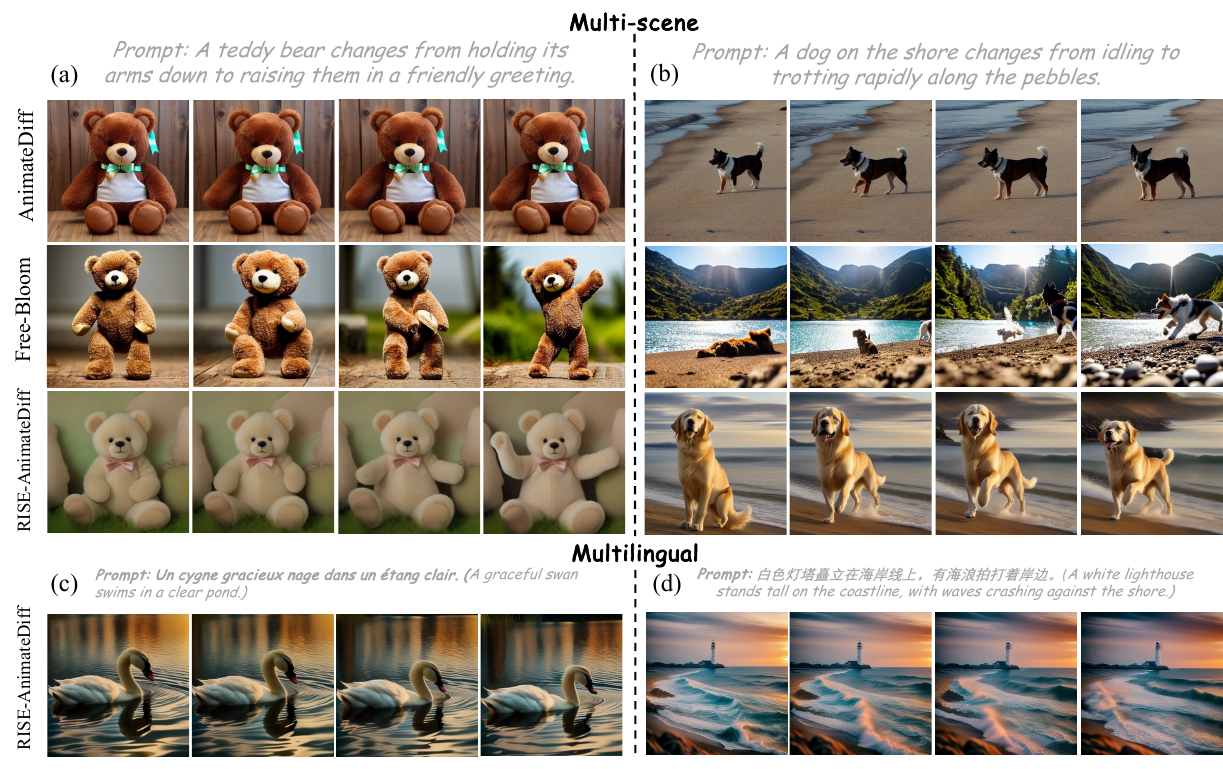}
    \vspace{-1em}
    \caption{Qualitative comparison on multi-scene and multilingual text encoding. In multi-scene video generation, the videos demonstrate high temporal alignment with the prompts. In multilingual video generation, we employ a French prompt in (c) and a Chinese prompt in (d).\vspace{-0.5em}}
    \label{fig:multi_scene}
    \vspace{-1em}
\end{figure*}
\begin{figure}[ht]
    \centering
    \includegraphics[width=0.9\linewidth]{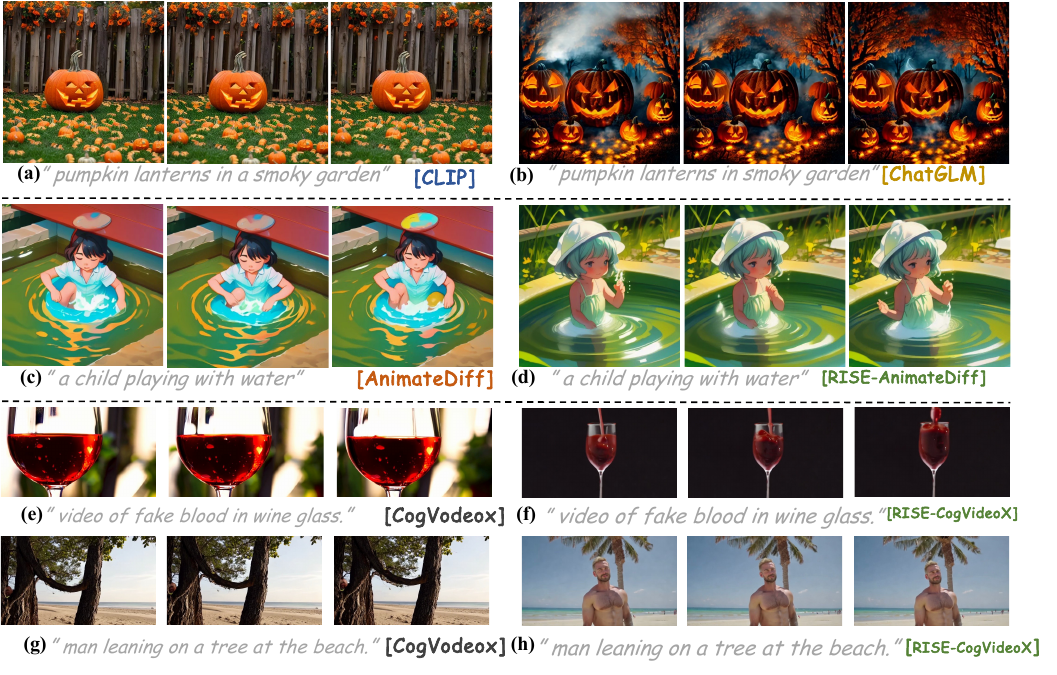}
    \vspace{-1em}
    \caption{(a). Utilization of CLIP as the text encoder. (b). Utilization of ChatGLM3 as the LLM. (c). Anime-style of AnimateDiff.  (d). Anime-style of RISE-AnimateDiff. (e)(g). Videos of CogVideoX. (f)(h). Video of RISE-CogVideoX\vspace{-0.5em}}
    \label{fig:ablation general}
    \vspace{-1em}
\end{figure}

\paragraph{The Analysis on Diverse LLMs and DMs.} 
By freezing the original weights of both the LLM and VDM throughout all stages of RISE-T2V training, we only need to align the RA's input dimensions with the rephrased hidden states and its output dimensions with the VDM's attention module. This design enables RISE-T2V to integrate seamlessly with various LLMs and VDMs. By replacing LLaMA2 with ChatGLM3~\cite{glm2024chatglm}, we integrated it into a UNet-based VDM. Figure \ref{fig:ablation general} (first row) shows the visualization and baseline comparisons. Switching to ChatGLM3 improved the video's meaning and aesthetics compared to the baseline.
We evaluate our method's performance with various Diffusion Models. We replaced the original Stable Diffusion weights with different style models from Civitai\footnote{https://civitai.com/models/30240/toonyou} \footnote{https://civitai.com/models/66347/rcnz-cartoon-3d} without additional training. As shown in Figure \ref{fig:ablation general} (second row), our method delivered satisfactory results. For Diffusion Transformer (DiT), we maintained LLaMA2 as the fixed LLM and integrated it with DiT model CogVideoX. Qualitative results are in the third and fourth rows of Figure \ref{fig:ablation general}. Our method under the DiT architecture produces results that are more aesthetically pleasing and better aligned with text than the baseline. Table~\ref{table:DMs} provides quantitative evaluations, demonstrating improvements over baseline models. These experiments validate the strong generalization capabilities of our approach.
\begin{table}[tbp]
    \centering 
    \setlength{\tabcolsep}{0.4mm}
    \footnotesize 
    \begin{tabular}{cccccc}
        \toprule
        Arch & Model & Domain &  Aesthetic$\uparrow$ & 
        Motion$\uparrow$ & 
        Text$\uparrow$ 
        \\

        \midrule
 			
        \multirow{4}*{UNet} & AnimateDiff & \multirow{2}*{2D Anime} & 7.04 & 0.986  & 28.44
        \\			
        & RISE-AnimateDiff &  & \textbf{7.37} & \textbf{0.992} & \textbf{30.13*}
        \\
        
        \cmidrule(lr){2-6}
        & AnimateDiff & \multirow{2}*{3D Cartoon} & 6.70 & 0.984 & 31.33
        \\			
        & RISE-AnimateDiff &  & \textbf{6.95} & \textbf{0.985} & \textbf{32.26*}
        \\
        \midrule
        \multirow{2}*{DiT} & CogVideoX & \multirow{2}*{-} & 5.40 & 0.983 & 30.12
        \\ 
        & RISE-CogVideoX &  & \textbf{5.43} & \textbf{0.995} & \textbf{33.33*} \\
        \bottomrule
    \end{tabular}
    \caption{\rm Ablation Study on Different Diffusion Models. }
    \vspace{-1em}
\label{table:DMs}
\end{table}
\begin{table}[t]
  \centering
  \footnotesize
    \begin{tabular}{cccc}
    \toprule
    Method & Aesthetic↑ & Motion↑ & Text↑ \\
    \midrule
    AnimateDiff & 6.39  & 0.983 & 31.36 \\
    AnimateDiff$_{ft}$ & 5.93  & 0.970 & 30.50 \\
    RISE-AnimateDiff & \textbf{6.61} & \textbf{0.984} & \textbf{31.95} \\
    \bottomrule
    \end{tabular}%
    \vspace{-0.5em}
    \caption{\rm Ablation Study on the Impact of Training Data. AnimateDiff$_{ft}$ is the baseline model fine-tuned on our exact same video dataset to ensure a fair comparison.\vspace{-0.6em}}
  \label{tab:data_impact_ablation}%
  \vspace{-1em}
\end{table}%

\paragraph{The Analysis on the Impact of Training Data.}
To rigorously verify that the performance improvement of our method comes from the proposed framework itself rather than simply benefiting from additional training data, we established a strong baseline, AnimateDiff$_{ft}$, by fine-tuning the original AnimateDiff model on exactly the same video dataset as used in our Stage 2 training. As shown in Table~\ref{tab:data_impact_ablation}, our RISE-AnimateDiff framework significantly outperforms AnimateDiff$_{ft}$ in terms of aesthetic quality, motion smoothness, and text alignment. Notably, simply adding our training data to AnimateDiff not only failed to bring improvement, but even led to performance degradation in certain metrics. These results strongly indicate that the superior performance of our method is not due to the training data.
\subsection{Video Generation with Multi-scene Text Encoding}
In terms of multi-scene video generation, we provide the LLM with a text prompt and instruct it to generate time-varying multi-scene text encodings. For better temporal alignment, each frame's feature map interacts with the corresponding text encoding in the attention module. We compared our approach with AnimateDiff and Free-Bloom~\cite{huang2024free}. Visual examples are presented in Figure~\ref{fig:multi_scene}. 
Specifically, in example (a), our method successfully captures the smooth motion of a teddy bear lowering and raising its hands. It shows superior text alignment, consistency, and aesthetics compared to AnimateDiff and Free-Bloom. 

\subsection{Video Generation with Multilingual Text Encoding}
We evaluate the ability of our method to generate videos from cross-language prompts despite all training texts being in English. During inference, a non-English prompt is entered into the LLM with instructions for rephrasing it in English, and the rephrased text encodings guide the video generation. As depicted in Figure~\ref{fig:multi_scene}, our approach creates videos with accurate semantics corresponding to the input prompts. These outcomes highlight the effectiveness of our approach in cross-language generation. By integrating the LLM, T2V models can be adapted to perform a wider variety of tasks through different rephrasing instructions.

\section{Limitations}
Despite the significant progress made, our work still has several limitations. The performance of our approach is constrained by the capabilities of the underlying video diffusion model. If the base model has insufficient spatial or temporal modeling abilities, the generated results may still suffer from motion discontinuity or limited visual quality, even with improved semantic conditioning. In future work, we will further evaluate and validate the effectiveness of our method on a wider variety of video diffusion models.

\section{Conclusion}
In this paper, we present RISE-T2V, which rephrases the text prompt and extracts semantic features in a seamless step. We propose the Rephrasing Adapter, a novel module that incorporates the LLMs as both a prompt rewriter and semantic feature extractor into T2V models. By employing the RISE-T2V framework, we leverage the powerful capabilities of LLMs to enable video generation models to accomplish a broader range of T2V tasks. Experimental results validate the effectiveness of our approach in three tasks. By modifying user instructions, RISE-T2V not only addresses the tasks outlined in this paper but also broadens potential applications. We aim to explore these opportunities further in future work.

\bibliography{aaai2026}

\newpage
\maketitle
\section{Appendix}
In the appendix, we begin by presenting the implementation details of our method and the details of dataset. Next, we outline the specifics of the experimental setup. We then present supplementary experimental results that validate our method. Finally, we show more visual results of the of our method. 
\section{Additional Details}

\subsection{Implementation Details}
We utilize the AdamW optimizer \cite{loshchilov2019decoupledweightdecayregularization} with a weight decay of 0.01 and maintain a consistent learning rate of 5e-5 for the first stage of RISE-T2V. For the second stage, the learning rate increases to 1e-4. The Video Diffusion Model (VDM) is configured
with a LoRA rank of 128. A noteworthy issue is that our training dataset for video often demonstrates lower visual quality compared to image datasets, primarily due to challenges encountered during collection. To mitigate the impact of this quality disparity on our temporal layers, while preserving the expertise ingrained in the pretrained spatial layers, we implement a strategy inspired by AnimateDiff. Specifically, during the second training stage, we use the rephrasing adapter that are finetuned using sampled static frames from video datasets. However, at inference time, we utilize the RA trained on image data from stage one.

All experiments are performed utilizing A100 GPUs, each equipped with 80GB of memory.
\subsection{Dataset Details}
\begin{figure}[htbp]
    \centering
    \includegraphics[width=1\linewidth]{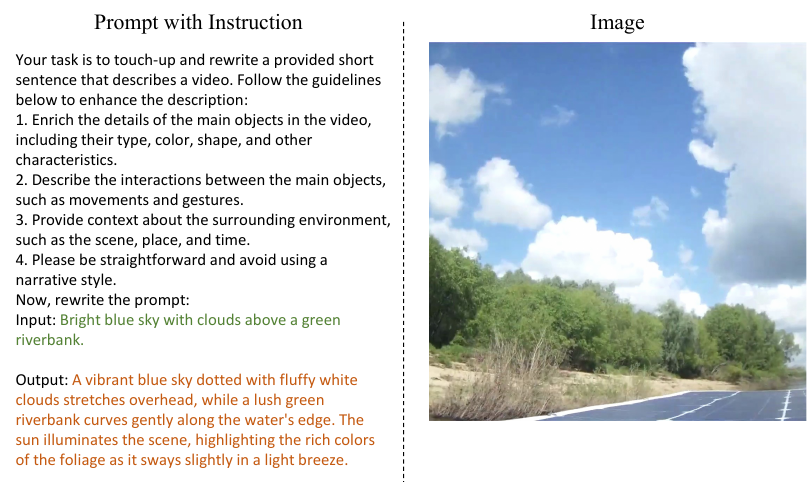} 
    \caption{An example from the stage 1 dataset. In this example, the black text indicates the user instruction, the green text represents the original caption of the video, and the red text shows the detailed caption generated after recaptioning.}
    \label{fig:sup_chatdata}
\end{figure}

\paragraph{The Training Data of Rephrasing Adapter} 
As mentioned in our main paper, we trained stage 1 using an internal dataset composed of 12 million text-image pairs with user instructions, which feature precise and detailed captions. Examples from the stage 1 dataset are illustrated in Figure \ref{fig:sup_chatdata}. Notably, the detailed captions were generated after utilizing a multimodal large language model \cite{bai2023qwenvlversatilevisionlanguagemodel} for recaptioning. The training data for Stage 2 comprises 1 million high-quality text-video pairs.

\section{Experiment Setup}
\subsection{Inference.}
Unless stated otherwise, we use Stable Diffusion with the Civitai\footnote{https://civitai.com/models/4201} checkpoint for our pre-trained spatial layers to align with the baseline methods in our experimental inference results. The DDIM \cite{song2022denoisingdiffusionimplicitmodels} sampler is employed, with the number of timesteps set to 25 and the classifier-free guidance scale \cite{ho2022classifierfreediffusionguidance} adjusted to 7.5.

\begin{figure}[htbp]
    \centering
    \includegraphics[width=1\linewidth]{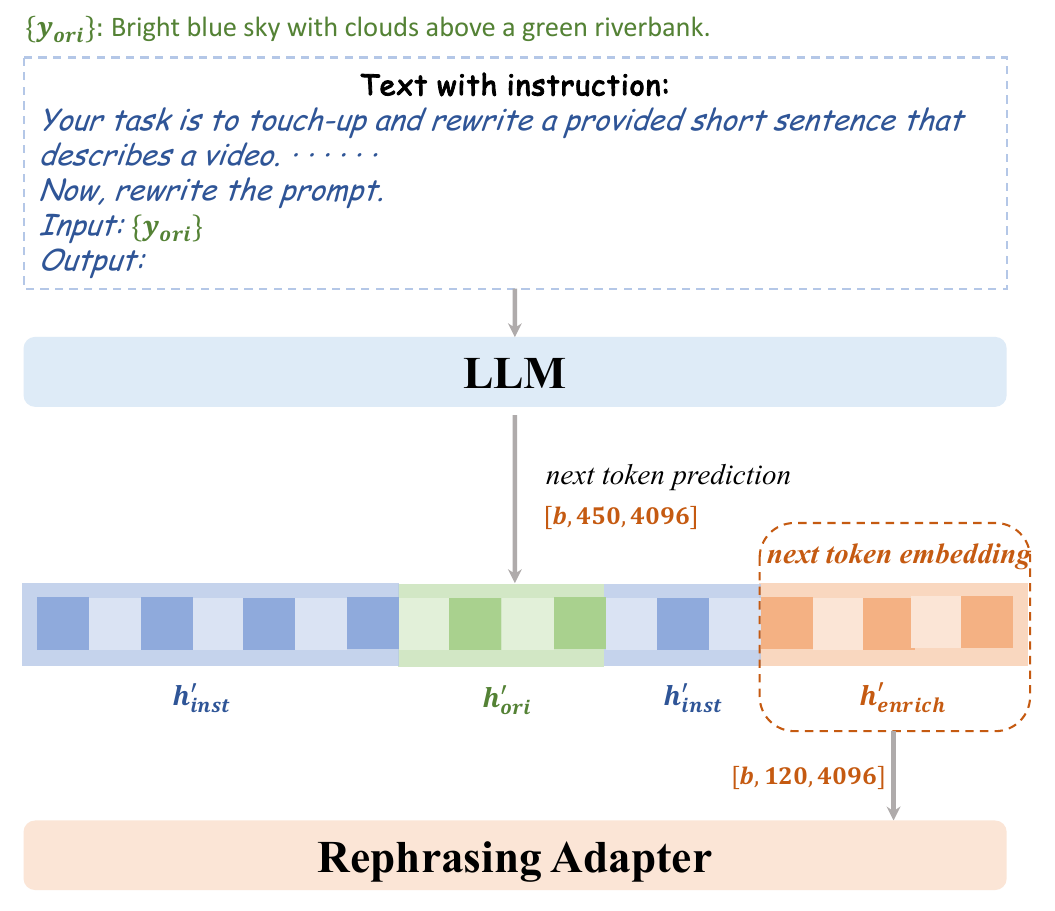} 
    \caption{An example of the processes involved in prompt rewriting and semantic feature extraction during inference.}
    \label{fig:sup_inference_hs}
    
\end{figure}

For dense text encoding generation, we performed a quantitative comparison on a random subset of VBench consisting of 800 prompts. These prompts are categorized into eight groups: animal, architecture, food, human, lifestyle, plant, scenery, and vehicles. Due to resource limitations, the test video data for the two commercial tools discussed in our main paper was provided by VBench. The user instructions used for testing are as follows:

\begin{itemize}
    \item \textit{Your task is to touch-up and rewrite a provided short sentence that describes a video. Follow the guidelines below to enhance the description:}
    
    \textit{Enrich the details of the main objects in the video, including their type, color, shape, and other characteristics.
    Describe the interactions between the main objects, such as movements and gestures.}
    
    \textit{Provide context about the surrounding environment, such as the scene, place, and time.}
    
    \textit{Use only English characters and punctuation.}
    
    \textit{Please be straightforward and avoid using a narrative style. Focus solely on the content of the video without describing the atmosphere or feelings.}
    
    \textit{Examples:}
    \textit{Input: The video shows a cup on a table.}
    
    \textit{Output: The video depicts a white ceramic coffee cup with a curved handle, positioned centrally on a wooden table with a textured surface, seemingly motionless, in a well-lit kitchen setting during the daytime.}
    
    \textit{Input: The video displays a cute cat.}
    
    \textit{Output: The video features a small, fluffy white cat with distinctive patches of ginger fur over its ears and back. The feline, adorned with a light-blue collar, is seen actively batting at a dangling feather toy with its paws, surrounded by a cozy living room setting with soft afternoon light filtering in.}
    
    \textit{Now, rewrite the prompt:}
    
    \textit{Input:}``\{$y_{ori}$\}''
    
    \textit{Output:}

\end{itemize}
In this context, $y_{\text{ori}}$ represents the original simple prompt input. To clarify which specific portions of the hidden states were utilized to generate videos, we provide a detailed explanation alongside the preceding user instructions and the examples illustrated in Figure \ref{fig:sup_inference_hs}. Let $ h' \in \mathbb{R}^{b \times s \times c} $ denote the total hidden states from the last layer of the LLM, obtained through next token prediction. Here, $ b $, $ s $, and $ c $ refer to the batch size, sequence length, and dimension size, respectively. Taking $ y_{\text{ori}} $ from Figure \ref{fig:sup_inference_hs} as an example, if the sequence length is 450, then $ h'_{inst} \in \mathbb{R}^{b \times 315 \times c} $, $ h'_{ori} \in \mathbb{R}^{b \times 15 \times c} $, and $ h'_{enrich} \in \mathbb{R}^{b \times 120 \times c} $. Thus, the next token embedding $ h'_{enrich} \in \mathbb{R}^{b \times 120 \times c} $ serves as the text encoding that guides the diffusion model in video generation.

For multi-scene text encoding generation, the test set comprises videos from the Free-Bloom dataset and some that we created ourselves. The user instructions utilized for testing this multi-scene video generation are as follows:
\begin{itemize}
    \item \textit{You will be provided with a video title. Your task is to describe the first and last frames of the video as if you are directing a movie.Please adhere to the following requirements:}
    
    \textit{1. Provide two descriptions: describe the start frame and the end frame of the video.}
    
    \textit{2. Use your imagination to create a continuous and varied narrative that matches the video content.}
    
    \textit{3. Each description should detail the subject's appearance and actions, outlining the main actions and the extent of actions.}
    
    \textit{4. Explicitly state attributes that remain unchanged between the beginning and end. Do not use pronouns like “the” to refer to these attributes.}

    \textit{5. Format your response as follows:
    detailed description related to the first frame|detailed description related to the last frame}
    
    \textit{Examples:}
    
    \textit{Example 1:}
    
    \textit{Video about: ``Volcano eruption''}
    
    \textit{Output: A towering volcano rises majestically amidst a vast stretch of undisturbed land, its peak piercing a backdrop of clear blue skies, and despite its imposing presence, there are no visible signs of smoke, ash, or any volcanic activity. \textbar~A towering volcano has roared to life once dormant and silent against the calm skies, now at the peak of its eruption, vehemently unleashing a searing pyroclastic flow that cascades fiercely down its slopes, engulfing the landscape in its fiery path.}
    
    \textit{Example 2:}
    
    \textit{Video about: ``A dachshund jumps up''}
    
    \textit{Output: A charming black and tan dachshund, clad in a whimsical red hat, stands alert on a shore blanketed with smooth, multi-colored pebbles, its curious eyes scanning the tranquil horizon as gentle waves lap at its stubby paws. \textbar~A charming black and tan dachshund with a bright red hat perched playfully atop its head bounds with sheer excitement along a beach, its paws scattering pebbles as it playfully leaps up and down the stone-strewn lakeside.}
    
    \textit{Now please provide two descriptions}
    
    \textit{Video about:} ``\{$y_{ori}$\}''
    
    \textit{Output:}

\end{itemize}

For multilingual text encoding generation, the user instructions utilized are as follows:
\begin{itemize}
    \item \textit{Role}
    
    \textit{You are a translator who can translate French into English while maintaining the original style and tone. You need to accurately convey information and cultural connotations, avoid literal translations, and focus on the expressive effect in the target language.}

    \textit{Skills}
    \textit{1. Translation: You need to translate the input text from the source language to the target language, ensuring the accuracy and fluency of the translation.
    2. Maintaining Style and Tone: You need to preserve the original style and tone as much as possible, including using the same vocabulary, sentence structures, and rhetorical devices.
    3. Conveying Cultural Connotations: You need to understand cultural differences between the source and target languages to ensure the translated content accurately conveys the original cultural nuances.
    4. Avoiding Literal Translation: You need to avoid literal translations, especially when dealing with content with significant cultural differences. You should translate according to the expression habits of the target language to ensure the translated content is easy to understand.}
    
    \textit{Constraints}
    
    \textit{- You can only translate textual content and not answer other questions.}
    
    \textit{- You need to use the language provided by the user to respond.}
    
    \textit{- You must follow the given format for responses and not deviate from the framework requirements.}
    
    \textit{- Please provide the translation directly without any additional responses.}
    
    \textit{Examples}
    \textit{Input: Un adorable chiot est sur le sol.
    Output: The video shows an adorable dog on the floor.</s>}
    
    \textit{Now, translate the input text into English.}
    
    \textit{Input: {caption}}
    
    \textit{Output:}
\end{itemize}

\subsection{Metrics.}
We employed four key evaluation metrics from VBench. The details are outlined below:
\paragraph{Aesthetic Quality} We assess the artistic and aesthetic value of each video frame as perceived by humans using the LAION aesthetic predictor \cite{aestheticpredictor}. This tool captures various aesthetic aspects, including layout, color richness and harmony, photo-realism, naturalness, and overall artistic quality. 
\paragraph{Motion Smoothness} 
It is essential to assess the smoothness of motion in generated videos and ensure it aligns with the physical laws of the real world. To evaluate this smoothness, we employ motion priors from a video frame interpolation model \cite{li2023amtallpairsmultifieldtransforms}.
\paragraph{Text alignment} To evaluate text alignment, we calculated the average similarity between the CLIP \cite{radford2021learning} prompt embedding and the CLIP embeddings of each individual frame.

We also computed the total score by calculating the average rank. This involves ranking all methods across the four aspects mentioned earlier and then determining the average.

\subsection{User Study.}
This paper presents three metrics requiring manual evaluation. Participants review four samples generated by different methods simultaneously and select the best one(s) based on three specific criteria: Aesthetic Quality, Temporal Quality, and Text Alignment. For each prompt, participants respond to the following questions:
\begin{itemize}
    \item Which sample exhibits the best overall performance in visual effects, composition, and color use in the video? (Aesthetic Quality)
    \item Which sample achieves the highest quality regarding the temporal scale of the video, considering the smoothness of motion and the dynamics' intensity? (Temporal Quality)
    \item Which sample demonstrates the greatest consistency between the video content and the textual prompt? (Text Alignment)
\end{itemize}
We randomly presented videos generated by our method alongside those produced by other methods to the participants. This process yielded a total of 300 valid results for dense text encoding generation.

\section{Supplemental Experiments}

\subsection{The Ablation on Diverse LLMs.}
We explore the incorporation of alternative open-source LLMs into our method, specifically ChatGLM3, which comprises 6 billion parameters. In our approach, we substitute LLaMA2 with ChatGLM3 and train RISE-AnimateDiff using the same training scheme and dataset. The automatic metric results, in comparison to AnimateDiff, are presented on the VBench test set as shown in Table \ref{table:chatglm}.
 
\begin{table}[htbp]
    \renewcommand{\arraystretch}{1}
    \centering
    \setlength{\tabcolsep}{0.4mm}
    \begin{footnotesize}
    \begin{tabularx}{\linewidth}{ccccc}
        \toprule
        Model &  Aesthetic$\uparrow$ & 
        Motion$\uparrow$ & 
        Dynamic$\uparrow$ & 
        Text$\uparrow$ 
        \\

        \midrule

        AnimateDiff & 6.39 & 0.983 & 0.083 & 31.36
        \\
        RISE-T2V(LLaMA2) & 6.61 & 0.984 & \textbf{0.273} &  \textbf{31.95}
        \\   
        RISE-T2V(ChatGLM3) & \textbf{6.72} & \textbf{0.985} & 0.151 & 30.90
        \\		
        \bottomrule
    \end{tabularx}
    \end{footnotesize}
    \caption{\rm Ablation study on the different LLM.}
\label{table:chatglm}
\end{table}
\section{More Visual Results}
\begin{figure*}[htbp]
    \centering
    \includegraphics[width=1\textwidth]{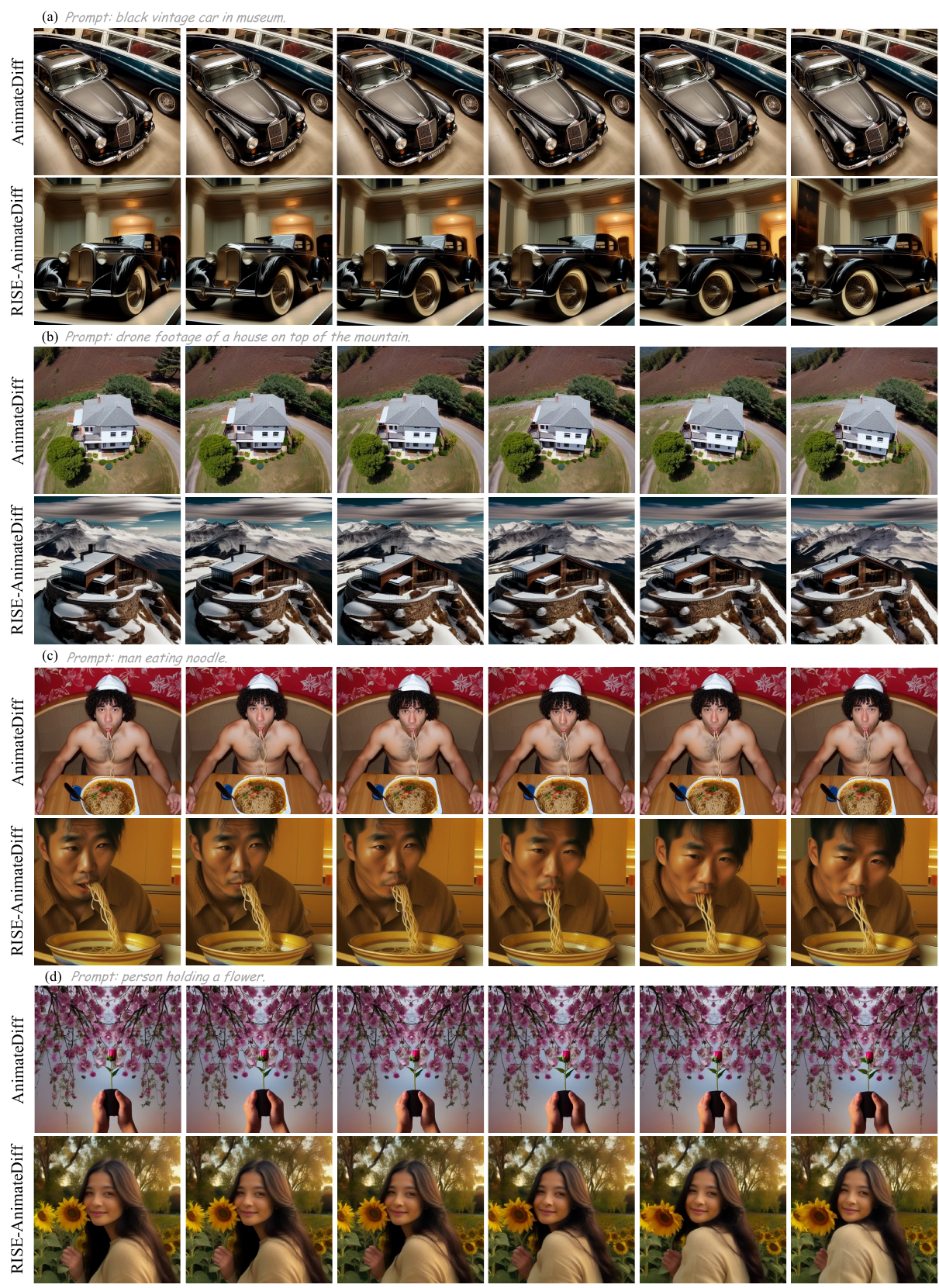} 
    \caption{Qualitative comparison on dense text encoding generation.}
    \label{fig:sup_lck}
\end{figure*}

\begin{figure*}[htbp]
    \centering
    \includegraphics[width=1\textwidth]{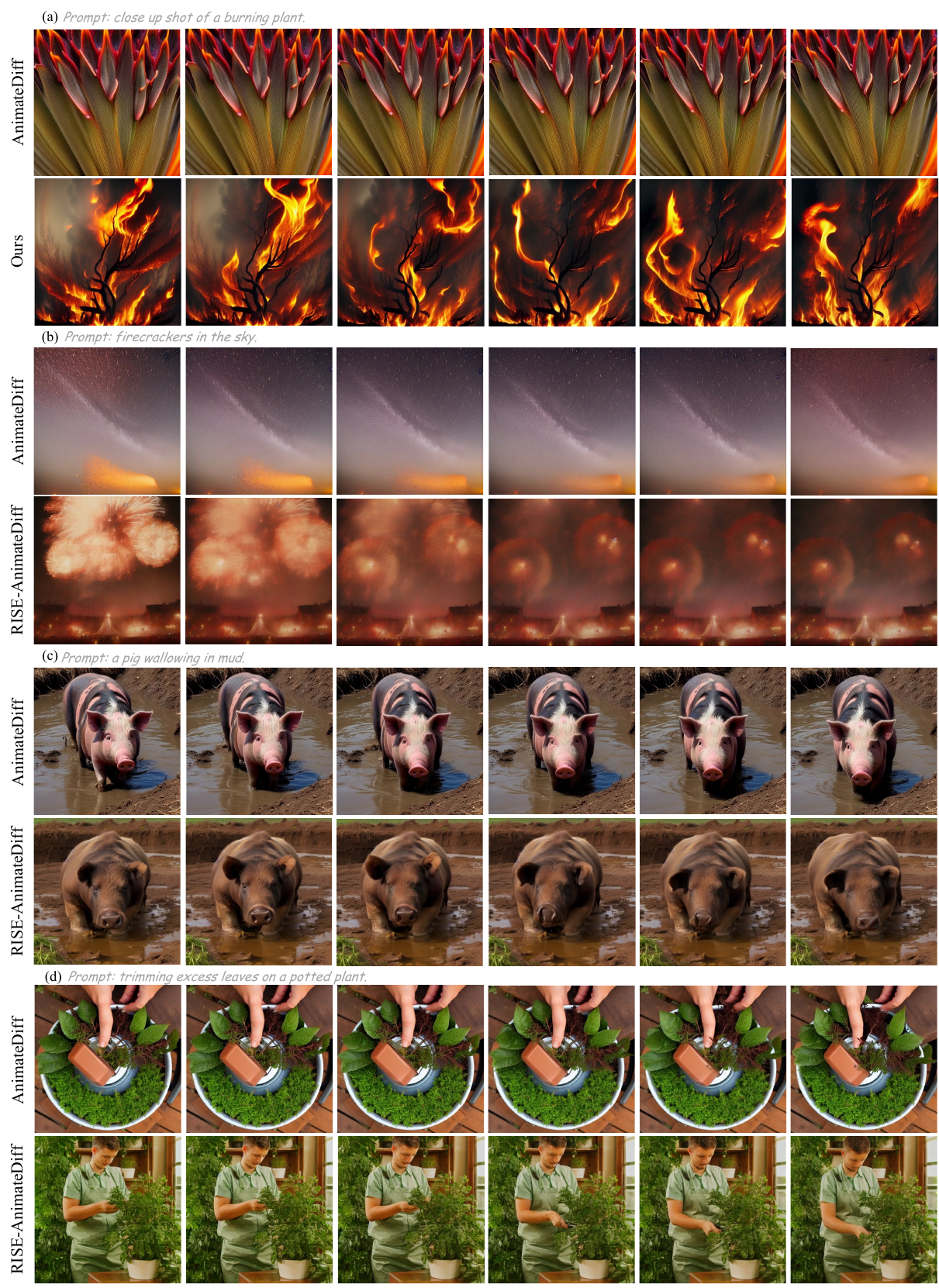} 
    \caption{Qualitative comparison of our method that utilizes ChatGLM3}
    \label{fig:sup_chatglm}
\end{figure*}

\begin{figure*}[htbp]
    \centering
    \includegraphics[width=0.88\textwidth]{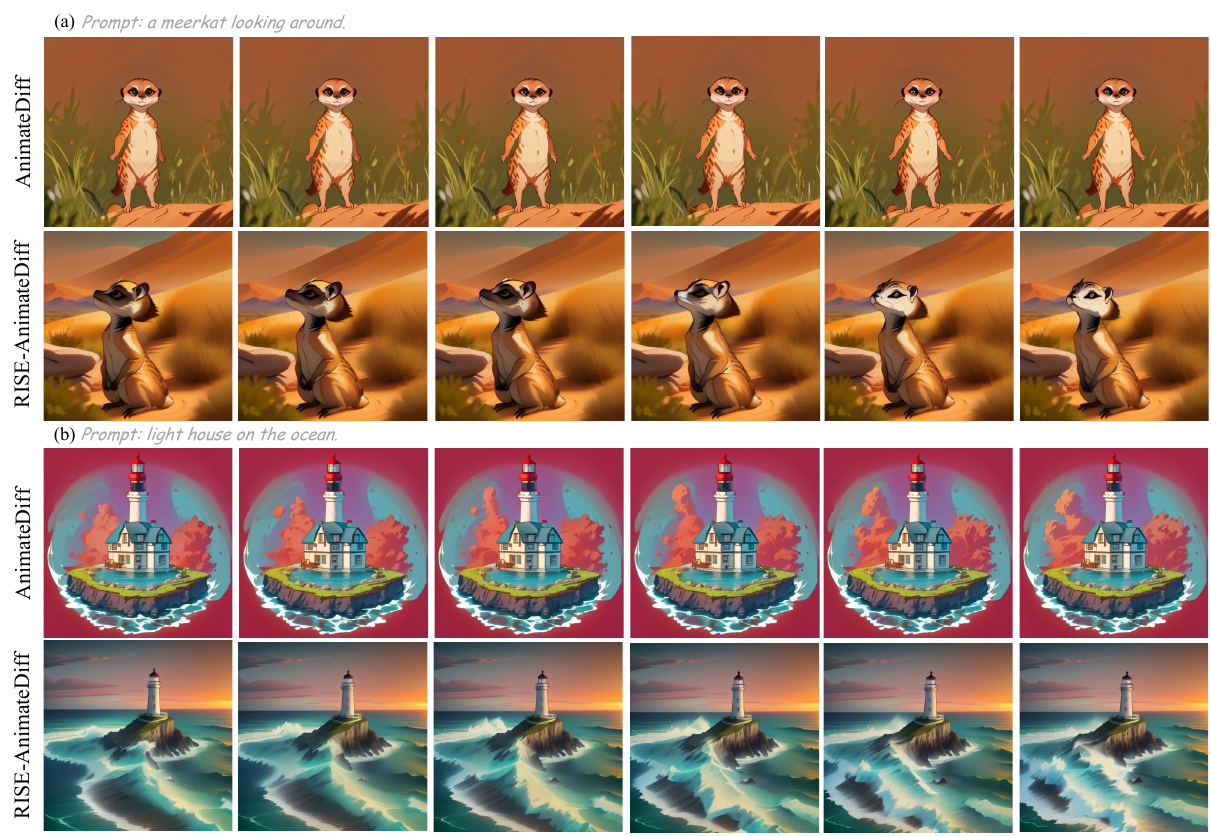} 
    \caption{Qualitative comparison of the 2D anime-style.}
    \label{fig:sup_toonyou}
    \vspace{-1em}
\end{figure*}

\begin{figure*}[htbp]
    \centering
    \includegraphics[width=0.88\textwidth]{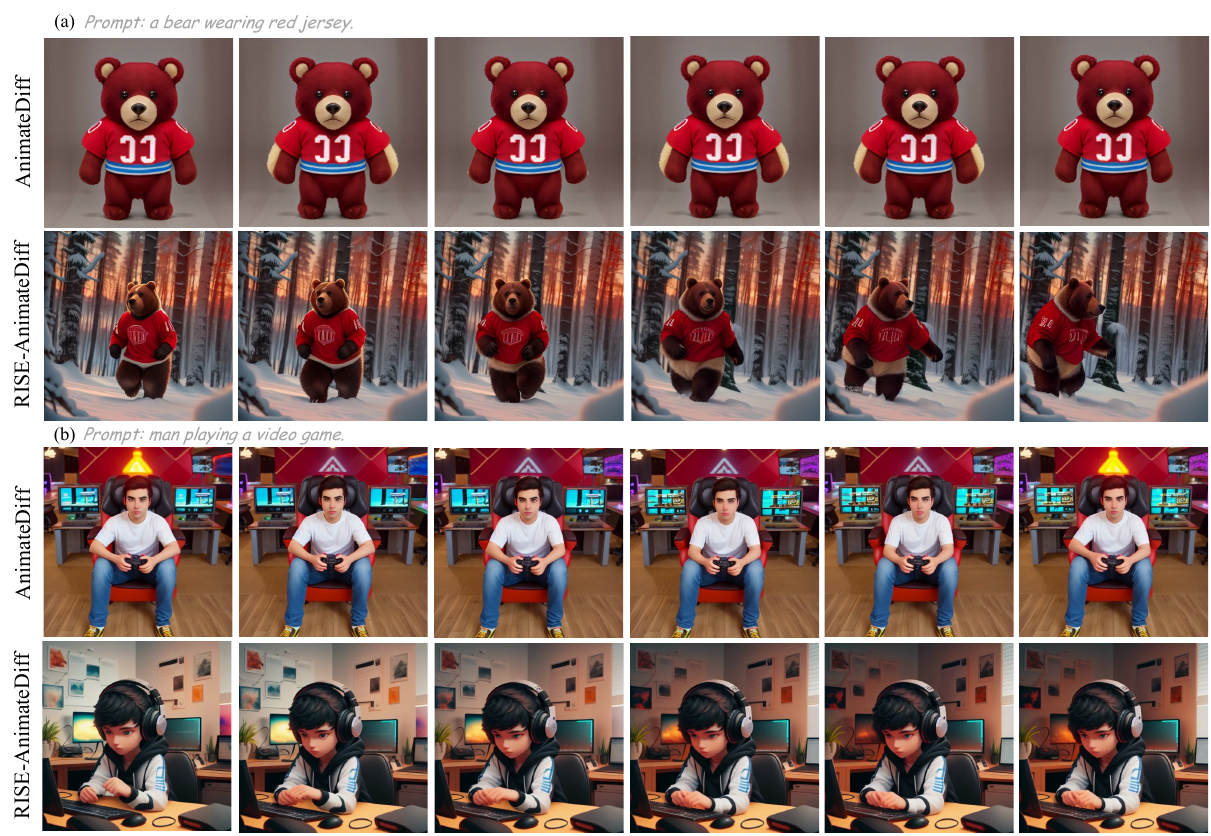} 
    \caption{Qualitative comparison of the 3D cartoon-style.}
    \label{fig:sup_3d}
    \vspace{-1em}
\end{figure*}

\begin{figure*}[htbp]
    \centering
    \includegraphics[width=1\textwidth]{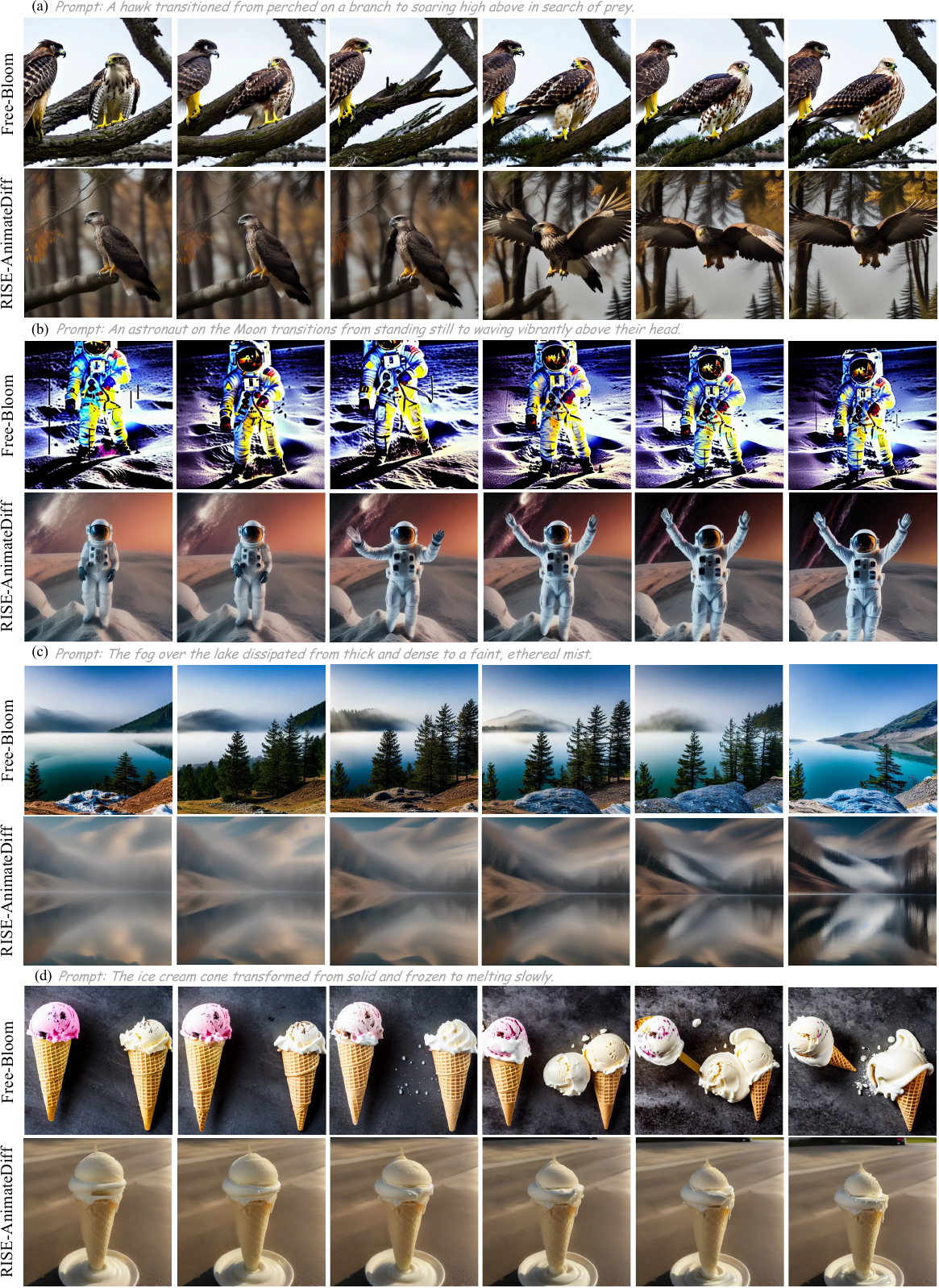} 
    \caption{Qualitative comparison on multi-scene text encoding generation.}
    \label{fig:sup_msv}
\end{figure*}

\begin{figure*}[htbp]
    \centering
    \includegraphics[width=1\textwidth]{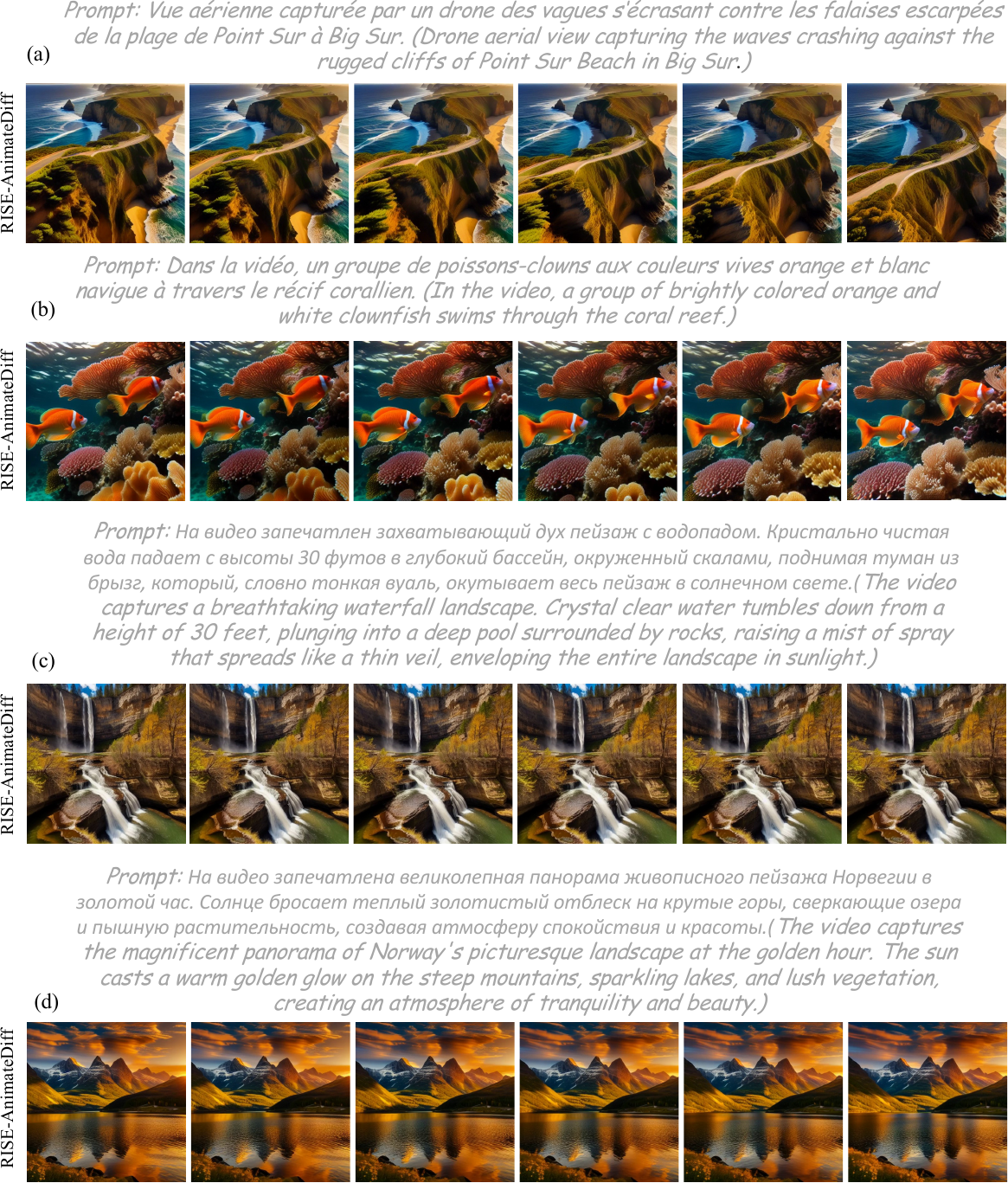} 
    \caption{Qualitative comparison on multilingual text encoding generation.}
    \label{fig:sup_ling}
\end{figure*}

In Figure \ref{fig:sup_lck}, we show more visual results of the dense text encoding generation. In Figure \ref{fig:sup_chatglm}, we showcase more visualization of our method equipped with ChatGLM3. In Figure \ref{fig:sup_toonyou} and \ref{fig:sup_3d}, We generated videos corresponding to both 2D anime-style and 3D cartoon-style. In Figure \ref{fig:sup_ling}, we show more qualitative comparisons on multilingual text encoding. In Figure \ref{fig:sup_msv}, we show more qualitative comparisons between our method and Free-Bloom for multi-scene. 





\end{document}